\documentclass[10pt,twocolumn,letterpaper]{article}

\usepackage{iccv}
\usepackage{times}
\usepackage{epsfig}
\usepackage{epstopdf}
\usepackage{graphicx}
\usepackage{amsmath}
\usepackage{amssymb}
\usepackage{tabularx}
\usepackage{multirow}

\usepackage[pagebackref=true,breaklinks=true,letterpaper=true,colorlinks,bookmarks=false]{hyperref}

\iccvfinalcopy 

\def\iccvPaperID{0000} 

\ificcvfinal\fi

\begin{document}

\newcommand{\prg}[1]{\vspace{0.1cm}\noindent\textbf{#1}~~}
\newcommand{\tabcapbefore}{\vspace{0.2cm}}
\newcommand{\tabcapafter}{\vspace{-0.2cm}}

\title{Second-order Convolutional Neural Networks\thanks{This research is funded by the Swiss National Science Foundation.}}

\author{Kaicheng Yu, Mathieu Salzmann\\
CVLab, EPFL\\
1015 Lausanne, Switzerland\\
{\tt\small \{kaicheng.yu, mathieu.salzmann\}@epfl.ch}
}

\maketitle

\begin{abstract}
\vspace{-0.4cm}
Convolutional Neural Networks (CNNs) have been successfully applied to many computer vision tasks, such as image classification. By performing linear combinations and element-wise nonlinear operations, these networks can be thought of as extracting solely first-order information from an input image. In the past, however, second-order statistics computed from handcrafted features, e.g., covariances, have proven highly effective in diverse recognition tasks. In this paper, we introduce a novel class of CNNs that exploit second-order statistics. To this end, we design a series of new layers that (i) extract a covariance matrix from convolutional activations, (ii) compute a parametric, second-order transformation of a matrix, and (iii) perform a parametric vectorization of a matrix. These operations can be assembled to form a Covariance Descriptor Unit (CDU), which replaces the fully-connected layers of standard CNNs. Our experiments demonstrate the benefits of our new architecture, which outperform the first-order CNNs, while relying on up to 90\% fewer parameters.

\end{abstract}

\vspace{-0.4cm}
\section{Introduction}

\begin{figure}[t]
\begin{center}
   \includegraphics[width=0.9\linewidth]{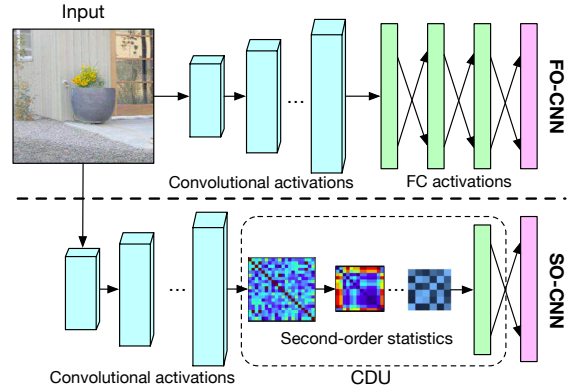}
\end{center}
\vspace{-0.5cm}
   \caption{{\bf Comparison of traditional first-order (FO-) CNNs~(top) with our second-order (SO-) CNNs~(bottom).} While, by performing linear combinations, traditional CNNs extract first-order information, our new architectures compute second-order statistics.}
   \label{fig:1v2}
   \vspace{-0.5cm}
\end{figure}

Image classification, e.g., recognizing objects and people in images, has been one of the fundamental goals of computer vision since its inception. In the past few years, Convolutional Neural Networks (CNNs), which jointly learn the features and the classifier, have proven highly effective at tackling such classification tasks~\cite{ Bell:2015ii,He:2015tt, Simonyan:2014ws}, and have thus dramatically accelerated the advances in recognition.
In essence, CNNs stack multiple layers, convolutional and fully-connected ones, with the parameters of each layer acting as filters on the output of the preceding one.
By computing such linear combinations, even when followed by element-wise nonlinearities and pooling, traditional CNNs can be thought of as extracting only first-order statistics from the input images. In other words, such networks cannot extract second-order statistics, such as covariances.

Psychophysics research, however, has shown that second-order statistics play an important role in the human visual recognition process~\cite{Julesz:1973ft}. This has been exploited in the past in computer vision via the development of Region Covariance Descriptors (RCDs)~\cite{Tuzel:2006vb}, which encode covariance matrices computed from local image features. In fact, these descriptors have been shown to typically outperform first-order features for visual recognition tasks such as material recognition and people re-identification~\cite{Harandi:2014hy, Harandi:2014hya, Jayasumana:2013we}. However, to this date, RCDs have been mostly confined to exploiting handcrafted features, and have thus been unable to match the performance of deep networks.

In this paper, we introduce a new class of CNN architectures that exploit second-order statistics for visual recognition. To this end, we develop three new types of layers. The first one extracts a covariance matrix from convolutional activations. The second one computes a parametric second-order transformation of an input matrix, such as a covariance matrix. Finally, the last one performs a parametric vectorization of an input matrix. These different types of layers can be stacked into a Covariance Descriptor Unit (CDU), which, as shown in  Fig.~\ref{fig:1v2}, replaces the fully-connected layers of a traditional CNN. Altogether, this provides us with second-order CNNs (SO-CNNs) that can be trained in an end-to-end manner.

To the best of our knowledge, only very few works have considered the use of RCDs in conjunction with CNNs. In particular,~\cite{Wang:2016va} extracted RCDs from features pre-computed using a CNN, but without proposing an end-to-end learning framework. By contrast,~\cite{Ionescu:2015wa} briefly studied the use of matrix outer product, which corresponds to a second-order operation, within a deep network as an application of their matrix backpropagation algorithm. While interesting, this work did not focus on extracting second-order statistics and thus remains preliminary in that respect. Here, we study this problem more thoroughly and introduce new layer types that were not considered in~\cite{Ionescu:2015wa}, and, as evidenced by our experiments, are key to the success of second-order CNNs.

We demonstrate the benefits of our second-order CNNs on the tasks of object recognition, using the CIFAR10 dataset \cite{krizhevsky2009learning}, and material recognition, using the challenging Materials in Context Database (MINC)~\cite{Bell:2015ii}. Our experiments demonstrate the generality of our approach by implementing it within different basic network architectures, such as FitNet~\cite{Romero:2014tg}, VGG16~\cite{Simonyan:2014ws} and ResNet~\cite{He:2015tt}. In all cases, we show that our second-order CNNs outperform the corresponding first-order ones, while relying on up to 90\% fewer parameters for networks having large fully-connected layers.
Furthermore, our method also outperforms the covariance learning framework of~\cite{huang2017}, using pre-computed deep features, and the single covariance network of~\cite{Ionescu:2015wa}. 
We believe that this clearly evidences the potential of our second-order CNNs and, by making our code publicly available, that it will motivate other researchers to explore going beyond first-order statistics within deep learning.

\vspace{-0.05cm}
\section{Related Work}
\vspace{-0.1cm}

Visual recognition is one of the core problems of computer vision, and has thus received a huge amount of attention. Below, we briefly review the recent advances that are most closely related to this work, which brings together the notions of deep learning and second-order statistics, such as covariance matrices.

\prg{CNNs for Visual Recognition.}
While, in the past, the problems of feature extraction and classifier training were typically decoupled~\cite{Calonder:2012vt,Lowe:2004kp,Schiele:2000ir}, the impressive results achieved 5 years ago by AlexNet~\cite{Krizhevsky:2012ju} on the ImageNet recognition challenge have put deep learning at the center of visual recognition. Recent years have seen great progress in this context, with increasingly deeper networks~\cite{ He:2015tt,Krizhevsky:2012ju, Simonyan:2014ws}, novel normalization~\cite{Ioffe:2015ud, Salimans:2016wo} and optimization~\cite{Duchi:2011kh,Kingma:2015us,Srivastava:2014ee,Zeiler:2012uw} strategies. All these networks, however, follow the same general strategy of stacking multiple layers, convolutional and fully-connected ones, each of which computes linear combinations of the output of the previous one. Despite the use of nonlinearities and pooling strategies, the resulting operations therefore still essentially extract first-order information, in the sense that they cannot compute higher-order statistics, such as covariances.

\prg{Covariance Descriptors for Visual Recognition.}
In the era of handcrafted features, however, second-order statistics, and particularly Region Covariance Descriptors (RCDs)~\cite{Tuzel:2008vu}, have proven effective to address visual recognition tasks. Several metrics have been proposed to compare RCDs~\cite{Arsigny:2006fx, Pennec:2006gp, Quang:2014uo, Sra:2012wh}, and they have been used in various classification frameworks, such as boosting~\cite{Tuzel:2008vu}, kernel Support Vector Machines~\cite{Jayasumana:2013we}, sparse coding~\cite{Cherian:2014bl, Guo:2010fm} and dictionary learning~\cite{Harandi:2015ko,Harandi:2012hp, Li:2013uh, Sra:2011gi}. In all these works, however, while the classifier was trained, no learning component was involved in the computation of the RCDs.

\prg{Covariance Descriptors and Learning.} 
To the best of our knowledge,~\cite{harandi:2014dr}, and its log-Euclidean metric learning extension~\cite{Huang:2015uu}, can be thought of as the first attempts to learn RCDs. 
This, however, was achieved by reducing the dimensionality of input RCDs, and thus has limited learning power. In a work concurrent to ours~\cite{huang2017}, the framework of~\cite{harandi:2014dr} was extended to learning multiple transformations of input RCDs. This approach, however, still relied on RCDs as input. By contrast, here, we introduce an end-to-end learning strategy. As discussed later, this requires special care to transition from the convolutional activations to the covariance matrix, and, as evidenced by our experiments, significantly outperforms the approach of~\cite{huang2017}.

\begin{figure*}[t]
\begin{center}
\includegraphics[width=\textwidth]{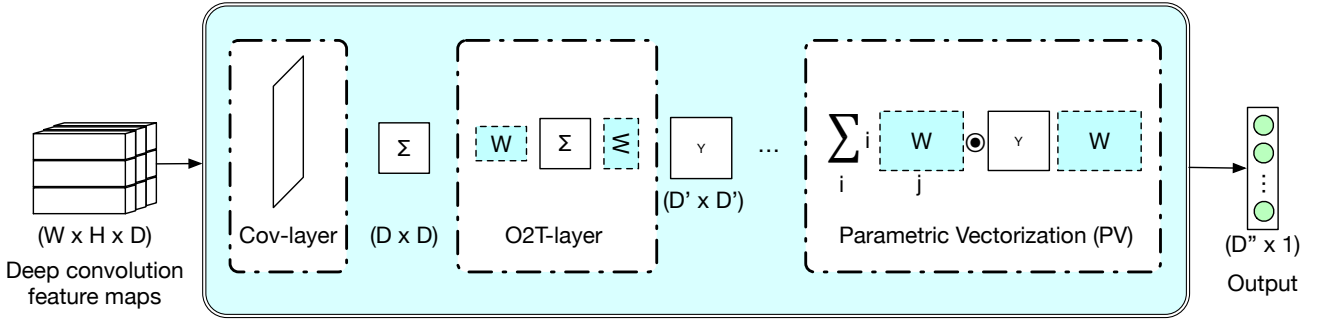}
\end{center}
   \caption{Our Covariance Descriptor Unit (CDU).}
\label{fig:dcov}
\end{figure*}

Only very few works have considered using RCDs in conjunction with deep learning. In particular,~\cite{Xu:2016kx} designed a CNN taking RCDs as input for the task of saliency computation. The focus of this work, however, differs fundamentally from ours, as it rather aims to process pre-computed RCDs, whereas we seek to learn second-order statistics from images. More closely related to our work,~\cite{Wang:2016va} computed RCDs from features extracted using a pre-trained CNN. Nevertheless, this work is limited to computing a standard covariance, and did not propose any end-to-end learning strategy. By contrast,~\cite{Ionescu:2015wa} briefly discussed the idea of computing a covariance matrix within a CNN, which was then flattened after a logarithmic map. Second-order statistics, however, were not the focus of this work, which rather aimed to develop a general matrix backpropagation algorithm. As a consequence, it did not consider practical problems such as the parameter explosion arising from appending a fully-connected layer to a large, flattened covariance matrix, and the resulting method would therefore not be applicable to networks with high-dimensional feature maps, such as the VGG or ResNet. Here, we not only take this into account, but also introduce new types of layers, thus truly developing a new class of deep architectures that exploit second-order statistics. Our experiments demonstrate that our second-order CNNs not only outperform the first order ones, but also the state-of-the-art covariance-based approaches of~\cite{Ionescu:2015wa} and~\cite{huang2017}.

\section{Our Approach}

In this section, we first introduce the basic architecture of our second-order CNNs (SO-CNNs), including our new layer types. We then address practical issues arising when starting from pre-trained convolutional layers and when dealing with high-dimensional convolutional feature maps.

\subsection{Basic SO-CNNs}
\label{sec:basic}

As illustrated by Fig.~\ref{fig:1v2}, an SO-CNN consists of a series of convolutions, followed by new second-order layers of different types, ending in a mapping to vector space, which then lets us predict a class label probability via a fully-connected layer and a softmax. The convolutional layers in our new SO-CNN architecture are standard ones, and we therefore focus the discussion on the new layer types that model second-order statistics. In particular, as illustrated by Fig.~\ref{fig:dcov}, we introduce three such new layer types: Cov layers, which compute a covariance matrix from convolutional activations; O2T layers, which compute a parametric second-order transformation of an input matrix; and PV layers, which perform a parametric mapping to vector space of an input matrix. Below, we discuss these different layer types in more detail.

\prg{Cov Layer.}
As suggested by the name, a Cov layer computes a covariance matrix. In particular, this type of layers typically follows a convolutional layer, and thus acts on convolutional activations.

Specifically, let $\mathbb{X}$ be the $(W \times H \times D)$ tensor corresponding to a convolutional activation map. This tensor can be reshaped into an $(N \times D)$ matrix $ {\bf X} = [ {\bf x}_1, {\bf x}_2, \dots, {\bf x}_N]$, with  ${\bf x}_k \in \mathbb{R}^D$ and $N=W\cdot H$. The $(D \times D)$ covariance matrix of such features can then be expressed as
\begin{equation} \label{eq:cov}
\boldsymbol{\Sigma} = \frac{1}{N} \sum_{k=1}^{N} ({\bf x}_k - \boldsymbol{\mu} )
({\bf x}_k - \boldsymbol{\mu} )^T \;,
\end{equation}
where $\boldsymbol{\mu} = \frac{1}{N} \sum_{k=1}^{N} \mathbf{x}_k$ is the mean of the feature vectors.

While $\boldsymbol{\Sigma}$ encodes second-order statistics, it completely discards the first-order ones, which may nonetheless bring valuable information. To keep the first order information, we propose to define the output of our Cov layer as
\begin{equation} \label{eq:cov_est}
{\bf C} =
\begin{bmatrix}
\boldsymbol{\Sigma} + \beta^2 \boldsymbol{\mu}\boldsymbol{\mu}^T
& \beta \boldsymbol{\mu} \\
\beta \boldsymbol{\mu}^T & 1 \\
\end{bmatrix} \;,
\end{equation}
which incorporates the mean of the features, via a parameter $\beta$. This parameter was set to $\beta = 0.3$ in our experiments.

A key ingredient for end-to-end learning is that the operation performed by each layer is differentiable. Being continuous algebraic operations, the covariance matrix in Eq.~\ref{eq:cov} and the mean vector $\boldsymbol{\mu}$ clearly are differentiable with respect to their input ${\bf X}$. This therefore makes our Cov layer differentiable, and enables its use in an end-to-end learning framework.

\prg{O2T Layer.}
The Cov layer described above is non-parametric. As a consequence, it may decrease the network capacity compared to the traditional way of exploiting the convolutional activations by passing them through a parametric fully-connected layer, and thus yield a less expressive model despite its use of second-order information. To overcome this, we introduce a parametric second-order transformation layer, which not only increases the model capacity via additional parameters, but also allows us to handle large convolutional feature maps.

More specifically, given a ($D \times D$) matrix ${\bf M}$ as input, our O2T layer performs a second-order transformation of the form
\begin{equation} \label{eq:o2t}
{\bf Y} = {\bf W} {\bf M} {\bf W}^T\;,
\end{equation}
whose parameters ${\bf W}  \in \mathbb{R}^{D \times D'} $ are trainable. Note that the value $D'$ controls the size of the output matrix, and thus gives more flexibility to the network than the previous Cov layer. Clearly, this second-order operation is differentiable, and can therefore be integrated in an end-to-end learning framework.

The O2T layer can be applied  either to a covariance matrix computed by a Cov layer, or recursively to the output of another O2T layer. Note that, since covariance matrices are symmetric positive (semi)definite (SPD) matrices, our formulation guarantees that the output obtained by applying one or multiple recursive O2T layers also is. To prevent degeneracies and guarantee that the rank of the original covariance matrix is preserved, additional orthonormality constraints can be enforced on the parameters ${\bf W}$. To this end, we make use of the optimization method on the Stiefel manifold employed in~\cite{Harandi:2016ug}. Empirically, we found these constraints to have varying but in general limited influence on the results. Altogether, our parametric O2T layers increase the capacity of the network while still modeling second-order information.

\prg{PV Layer.}
Since our ultimate goal is classification, we eventually need to map our second-order, matrix-based representation to a vector form, which can in turn be mapped to a class probability estimate via a fully-connected layer with a softmax activation. In~\cite{huang2017, Ionescu:2015wa}, such a vectorization was achieved by simply flattening the matrix, after applying a logarithmic map. When working with large matrices (large $D$), this however may lead to an intractable number of parameters to map the resulting $\mathcal{O}(D^2)$-dimensional vector to the vector of class probability estimates. Here, instead of direct flattening, we introduce a parametric vectorization of the second-order representation.

Specifically, given an input matrix ${\bf Y} \in \mathbb{R}^{D' \times D'}$, we compute a vector $\mathbf{v} \in \mathbb{R}^{D''}$, whose $j$-th element is defined as
\begin{equation}
[\mathbf{v}]_j = ([{\bf W}]_{:,j})^T {\bf Y} [{\bf W}]_{:,j}  = \sum_{i=1}^{D'}
[{\bf W} \odot {\bf Y} {\bf W}]_{i,j}\;,
\label{eq:wv}
\end{equation}
where ${\bf W} \in \mathbb{R}^{D' \times D''}$ are trainable parameters, and $[\mathbf{A}]_{i,j}$ denotes the entry in the $i$-th row and $j$-th column of matrix $\mathbf{A}$, with $[\mathbf{A}]_{:,j}$ the complete $j$-th column. Note that, while both formulations in Eq.~\ref{eq:wv} are equivalent, the first one is easier to interpret, but the second one is better suited for efficient implementation with matrix operations.

Due to its formulation, this vectorization can, in essence, still be thought of as a second-order transformation. More importantly, being parametric, it increases the flexibility of the model, while preventing the number of parameters in the following fully-connected layer to become intractable. As for our other layers, this operation is differentiable, and can thus be integrated to an end-to-end learning formalism.

\prg{General SO-CNN Architecture.}

We dub Covariance Descriptor Unit (CDU) a sub-network obtained by stacking our new layer types.
In short, and as illustrated in Fig.~\ref{fig:dcov}, a CDU takes as input the activations of a convolutional layer and first computes a covariance matrix according to Eq.~\ref{eq:cov_est}. The resulting matrix is passed through a number of O2T layers (Eq.~\ref{eq:o2t}), including none, whose output is then mapped back to a vector via a PV layer. Each of these layers can be followed by an element-wise nonlinearity. In particular, we make use of ReLUs, which have the property of maintaining the positive definiteness of SPD matrices. Importantly, the resulting CDUs are generic and can be integrated in any state-of-the-art CNN architecture.

As such, our framework makes it possible to transform any traditional first-order CNN architecture into a second-order one for image classification. To this end, one can simply remove the fully-connected layers of the first-order CNN and connect the resulting output to a CDU. The output of the CDU being a vector, one can then simply pass it to a fully-connected layer, which, after a softmax activation, produces class probabilities. Since, as discussed above, all our new layers are differentiable, the resulting network can be trained in an end-to-end manner.

\subsection{Starting from Pre-trained Convolutions}
\label{sec:transition}
The basic SO-CNN architecture described above can be trained from scratch, which we will show in our experiments. To speed up training, however, one might want to leverage the availability of pre-trained first-order CNNs. To do so, we propose to first freeze the pre-trained convolutional layers to train the second part of the SO-CNN, and then fine-tune the entire network. We observed empirically that, while we could train the second part of the network, fine-tuning did not converge. This, we believe, is due to the fact that there is no connection between first- and second-order features in the first stage, and thus the gradient of the second part is too different from that of the first one at the beginning of the fine-tuning process. 

To address this, we therefore propose to introduce an additional transition layer, which will facilitate training and give more flexibility to the model by allowing it to modify the pre-trained convolutional feature maps.

To this end, we apply a linear mapping to each feature vector independently. Specifically, let ${\bf x}_k$ be an original convolutional feature vector. We then learn a mapping of the form
\begin{equation}
h(\mathbf{x}_k) = \mathbf{Wx}_k + \mathbf{b}\;,
\end{equation}
where ${\bf W} \in \mathbb{R}^{\tilde{D}\times D}$ is a trainable weight matrix, and $\mathbf{b} \in \mathbb{R}^{\tilde{D}}$ a trainable bias. By constraining the weight matrix and bias to be the same for all the feature vectors, this is equivalent to a $1 \times 1$ convolutional layer with linear activation function. The parameter $\tilde{D}$ gives rise to a range of different models, with adapted features ranging from lower to higher dimensionalities than the original ones. As shown in our experiments, this strategy allows us to effectively exploit pre-trained convolutions in our SO-CNNs, while still learning the entire model in an end-to-end manner by unfreezing the convolutions in a second learning phase.

\begin{figure}[t]
\begin{center}
\vspace{-0.1cm}
   \includegraphics[width=1\linewidth]{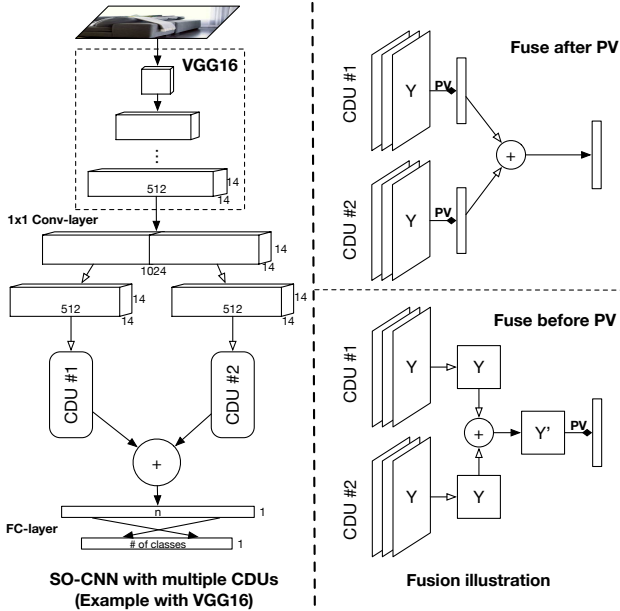}
\end{center}
   \caption{{\bf Using multiple CDUs.}
  {\bf (Left)} Example of an SO-CNN with multiple CDUs.
  {\bf (Right)} Two methods to fuse information between multiple CDUs. Fusion occurs after the PV-layers in the top figure, and before in the bottom one. Fusion strategies include concatenation, summation and averaging. 
  Note that black arrows indicate mathematical operations, whereas white ones correspond to an identity mapping.
  }
\label{fig:fuse}
\end{figure}
\vspace{-0.3cm}

\subsection{Handling High-dimensional Feature Maps}
\label{sec:robust}
In our basic SO-CNNs, a CDU directly follows a convolutional layer. While this transition can, in principle, be achieved seamlessly, the rapid growth in the dimensionality of the convolutional feature maps computed by modern architectures makes this problem more challenging. Indeed, with a basic architecture derived from, e.g., the ResNet~\cite{He:2015tt}, whose last convolutional activation map has size $(7 \times 7 \times 2048)$ for a $(224 \times 224)$ input, the resulting covariance matrix would be very high-dimensional ($2048 \times 2048$), but have a low rank (at most 48). In practice, this would translate into instabilities in the learning process due to many 0 eigenvalues. While, in principle, this could be handled by using the  strategy of Section~\ref{sec:transition} with a small $\tilde{D}$, this would incur a loss of information that reduces the network capacity too severely.  Below, we study two strategies to overcome this problem, which define our complete SO-CNN architecture.

\prg{Robust Covariance Estimation.}
As a first solution to overcome the low-rank problem, we make use of the robust covariance approximation introduced in~\cite{Wang:2016va} in the context of RCDs. Specifically, let $\boldsymbol{\Sigma} = {\bf U}  {\bf S} {\bf U}^T$ be the eigenvalue decomposition of the covariance matrix. A robust estimate of $\boldsymbol{\Sigma}$ can be written as
\begin{equation}
 \hat{ \boldsymbol{\Sigma} } = {\bf U} f({\bf S}) {\bf U}^T\;,
\end{equation}
where $f(\cdot)$ is applied element-wise to the values of the diagonal matrix ${\bf S}$, and is defined as
\begin{equation} \label{eq:robust}
f(x) = \sqrt{\bigg(\frac{1-2\alpha}{2\alpha}\bigg)^2 + \frac{x}{\alpha}} - \frac{1 - \alpha}{2\alpha}\;,
\end{equation}
with parameter $\alpha$ set to $0.75$ in practice. The resulting estimate $\hat{ \boldsymbol{\Sigma} }$ can then replace $\boldsymbol{\Sigma}$ in Eq.~\ref{eq:cov_est}.

Thanks to the matrix backpropagation framework of~\cite{Ionescu:2015wa}, which handles eigenvalue decomposition, this robust estimate can also be differentiated, and thus incorporated in an end-to-end learning framework.

\prg{Multiple CDUs.}
Our second strategy to handling high-dimensional feature maps, illustrated by Fig.~\ref{fig:fuse}(left), consists of splitting the feature maps into $n$ separate groups of equal sizes. Each group will then act as input to a different CDU, whose covariance matrix will have fewer 0 eigenvalues than a covariance obtained from all the features. For example, with a ResNet, instead of computing a covariance descriptor of size $2048 \times 2048$, we create 4 groups of 512 features, and use them to compute 4 different covariance descriptors, followed by separate O2T and PV layers. In essence, this strategy still makes use of all the features, but does not consider all the possible pairwise covariances. However, since the features are learned, the network can automatically determine which pairwise covariances are important. Note that the robust covariance estimate discussed above can be applied to the covariance matrix of each group.

Ultimately, the information contained in the multiple CDUs needs to be fused into a single image representation. We propose two strategies to do so, illustrated in Fig.~\ref{fig:fuse}(right). The first one consists of combining the CDUs output vectors by an operation such as summing, averaging or concatenation. The second one aims at fusing the multiple branches before vectorization, which can be again achieved by summing or averaging the respective matrices, or concatenating them into a larger block-diagonal matrix. This is then followed by a PV layer.

\section{Experiments}

In this section, we first present results obtained with our basic SO-CNN introduced in Section~\ref{sec:basic} on CIFAR-10. We then turn to evaluating our complete SO-CNN architecture, with the different strategies introduced in Sections~\ref{sec:transition} and~\ref{sec:robust}, on the larger, more challenging MINC dataset.

\subsection{Basic SO-CNNs on CIFAR-10}
\label{sec:cifar_result}
\textbf{CIFAR-10}~\cite{krizhevsky2009learning} is an object recognition dataset containing 50000 training and 10000 testing $ (32 \times 32)$ RGB images depicting 10 classes of objects. In the following experiments, we augmented the data by flipping the training images for all models and baselines. Because of the relatively small scale of this dataset, we can directly apply our basic SO-CNN to it. We therefore make use of this dataset to evaluate different architecture designs within our basic SO-CNN framework. Furthermore, we compare our basic SO-CNN to the corresponding first-order CNN, to the matrix backpropagation model of~\cite{Ionescu:2015wa} (\emph{MatBP}) and to the \emph{SPD-net} of~\cite{huang2017}.

\begin{figure}
\vspace{-0.2cm}
	\includegraphics[width=\columnwidth]{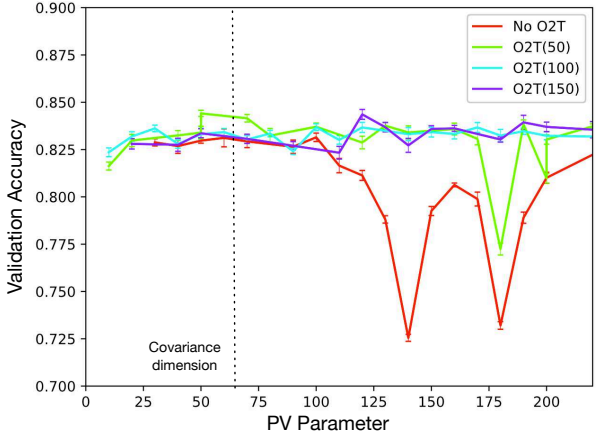}
    \caption{{\bf Joint influence of the PV output dimension and the second-order dimension.}
    With no O2T layers, learning is unstable once the PV dimension becomes significantly larger than the covariance dimension (64). With one O2T layer, learning is more stable, particularly when the PV dimension is not significantly larger than the O2T dimension. This suggests one should use a PV dimension similar to that of the last O2T layer.
    }
    \label{fig:wv_result}
    \vspace{-0.4cm}
\end{figure}

\prg{Model Setup.} We use the FitNet-v1 model of~\cite{Romero:2014tg} as our base first-order architecture. FitNet has 3 convolutional blocks, each of which contains 3 convolutional layers, with no dropout. 
The filters are of size $(3,3)$ for all layers, and one max-pooling layer is attached after each block. In the first-order model, the last convolutions are followed by one fully-connected (FC) layer of size 500. In our basic SO-CNNs, we replace this layer with a CDU. Since the last convolutional feature map is of dimension 64, the resulting covariance matrix is sufficiently small not to require a robust estimate or multiple CDUs. Both FitNet and our SO-CNN have then a final FC layer to produce a 10-dimensional vector of class probabilities via a softmax activation. Below, we evaluate different architectures of our SO-CNN model, corresponding to varying the output dimensionality of the PV layer, and the number and dimensionalities of the O2T layers. For all models (first- and second-order), all the weights were initialized using the method of~\cite{Glorot:2010uc}. We used stochastic gradient descent with an initial learning rate of 0.01 and reduced by a factor 10 when the validation loss does not decrease for 8 epochs.

\begin{table}[!t]
\vspace{-0.2cm}
 \centering
  \begin{tabular}[c]{l|l|l|l|l}
    \hline
    \textsc{Setting}  & \begin{footnotesize}SO-CNN-2 \end{footnotesize}& \begin{footnotesize}SO-CNN-3 \end{footnotesize}& \begin{footnotesize}SO-CNN-4 \end{footnotesize}& \begin{footnotesize}SO-CNN-5 \end{footnotesize} \\
    \hline
    \textsc{Same}      &    $82.90\%$    & $83.68\%$  & $83.18\%$  & $84.07\%$ \\
    \textsc{$\div 2$}  &    $82.86\%$    & $84.45\%$  & $83.69\%$   & $83.39\%$ \\
    \textsc{$\times 2$}&    $83.35\%$    & $84.77\%$  & $\boldsymbol{85.10\%}$  & $84.04\%$ \\ \hline
  \end{tabular}
  \tabcapbefore
  \vspace{-0.2cm}
  \caption{{\bf Influence of O2T layer number and dimension.} \textsc{Same} indicates that the dimension is the same (64) in all layers, and $\div 2$ or $\times 2$ that the dimension is divided or multiplied by 2 from one layer to the next. The PV-layer has the same dimension as the last O2T-layer. For example, CDU-3 with $\div 2$ corresponds to O2T(200) - O2T(100) - O2T(50) - PV(50).}
  \tabcapafter
  \label{tab:valid_o2t}
  \vspace{-0.6cm}
\end{table}

\prg{PV Output Dimension vs Second-order Dimension.} 
Intuitively, the output dimensionality of the PV layer should be similar to that of the second-order descriptor, whether the last O2T layer or directly the covariance when no O2T layer is used (e.g., a much smaller dimension would result in information loss). In a first experiment, we therefore evaluate the joint influence of these two dimensionalities. To this end, we make use of either no O2T layer,
or one such layer
We vary the PV output dimensionality from 10 to 200 with a step size of 10, and the dimensionality of the O2T layer, denoted by O2T($m$) for dimension $m \in \{50, 100,150\}$. In Fig.~\ref{fig:wv_result}, we plot the accuracy of the resulting models as a function of the PV dimensionality.  We can observe that a small $m$ should be used in conjunction with a small PV dimension, whereas a large $m$ yields slightly higher accuracy with a high PV dimension. Furthermore, training seems to be less stable if the PV dimension is significantly larger than the second-order one. We can also see that, as expected, adding one O2T layer brings more flexibility to the model, and thus yields higher accuracy.

\prg{Number and Dimensions of O2T Layers.}
As a second experiment, we evaluate the influence of the number and dimensions of O2T layers in our SO-CNN framework. To this end, we vary the number of O2T layers from 2 to 5 (we also tested with 1 but omit it here due to a consistently slightly lower accuracy), denoted by SO-CNN-\{2,3,4,5\}, and follow three strategies regarding their dimensionalities: (i) We keep the dimension constant across the different O2T layers; (ii) We increase the dimensionality from 50 by a factor 2 in successive O2T layers; (iii) We decrease the dimensionality by half to reach a final dimension of 50. In all these settings, following the results of the previous experiment, we set the PV output dimensionality to that of the last O2T layer.
The results of this experiment are provided in Table~\ref{tab:valid_o2t}. They show that (i) adding more O2T layers indeed increases the capacity of the network, but may lead to overfitting if too many such layers are employed; (ii) the most effective strategy to set the dimensionalities of the O2T layers consists of increasing them in successive layers.

\prg{Comparison to the Baselines.}
Following the previous analysis, in Table~\ref{tab:fitnet_result}, we compare the results of our SO-CNN-4 model with increasing O2T layer dimensions and PV output dimension matching that of the last O2T layer with the first-order FitNet CNN, and the MatBP~\cite{Ionescu:2015wa} and SDP-Net~\cite{huang2017} baselines. For the comparison to be fair, for MatBP, we made use of the same FitNet-based architecture as us. For SDP-net, which relies on a covariance matrix as input, we exploited RCDs obtained from the last convolutional layer of the first-order FitNet.  Note that we were unable to train these two baselines from scratch, as opposed to our SO-CNNs, and therefore fine-tuned them from the pre-trained FitNet. The hyper-parameters of SDP-net were set according to the recommendations in~\cite{huang2017}. 

\begin{table}[!t]
\vspace{-0.2cm}
 \centering
  \begin{tabular}[c]{l|l|l|l}
    \hline
    \textsc{Classifier} & \textsc{Settings} 
    &\textsc{$\#$ Params} & \textsc{Acc} \\
    \hline
    FitNet~\cite{Romero:2014tg} & 500 & 620K & $83.15\%$ \\
    MatBP~\cite{Ionescu:2015wa} & - & 131K & $28.27\% $ \\
    SPD-net~\cite{huang2017} & 70,50,30 & 55K & $76.07\%$ \\
    \hline
   	\hline
    SO-CNN-4 & $\times 2$ & 362K & $\boldsymbol{85.10\%}$ \\ \hline
  \end{tabular}
  \tabcapbefore
  \caption{{\bf Baseline comparison on CIFAR10 with FitNet architectures.} 
 Note that we outperform all baselines, while relying on roughly 40\% fewer parameters than the first-order CNN, which is closest to us in accuracy.}
  \tabcapafter
  \label{tab:fitnet_result}
  \vspace{-0.2cm}
\end{table}

As can be seen from the table, our model outperforms MatBP and SPD-net by a significant margin, thus showing the benefits of our end-to-end learning strategy over using a single covariance flattened after a log-map (MatBP) and over a two-stage strategy consisting of using a pre-defined covariance matrix as input (SDP-net). Note also that our model outperforms the first-order one, thus showing the importance of leveraging second-order information. As can be verified from the results of our previous experiments, other versions of our SO-CNN also outperform the first-order one, confirming the benefits of our approach. Altogether, we believe that these results clearly demonstrate the potential of our basic SO-CNN architecture.

\subsection{Complete SO-CNNs on MINC}
We now evaluate our compete SO-CNNs, including the strategies introduced in Sections~\ref{sec:transition} and~\ref{sec:robust}, on the large-scale MINC material recognition dataset. This choice was motivated by the fact that traditional second-order descriptors have proven particularly effective for tasks such as material or texture recognition~\cite{Harandi:2014hy, Harandi:2014hya, Jayasumana:2013we}. Below, we briefly describe this dataset and the architectures we used, as well as evaluate different versions of our approach and compare it to the state-of-the-art.

\begin{figure}[t]
\vspace{-0.2cm}
    \centering
    \includegraphics[width=0.9\linewidth]{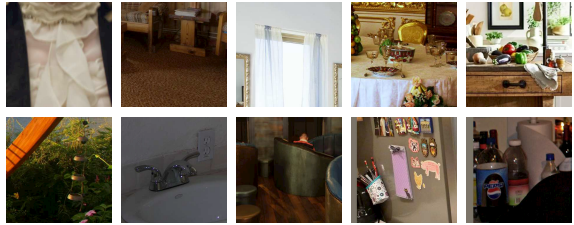}
    \caption{Samples from the MINC-2500 dataset}
    \label{fig:minc}
    \vspace{-0.4cm}
\end{figure}

{\bf MINC-2500} is a large-scale material recognition dataset containing 23 classes of different materials, some of which are shown in Fig.~\ref{fig:minc}. For each class, there are 2500 $(362 \times 362)$ RGB images. We split the dataset into training, validation and test samples with proportions $0.85, 0.05, 0.10$, respectively. Unlike other small-scale material databases~\cite{Caputo:2005gf,Cimpoi:2014vq, Sharan:2009jq}, the images contain not only the material but also its surrounding environment, thus making it more challenging. 
To augment the data, we used horizontal flip, and random cropping to $224 \times 224$ patches, thus matching the standard input size of our base CNN architectures described below.

\prg{CNN Architectures.}
The size of this dataset makes it well-suited to use recent deeper architectures, such as the VGG~\cite{Simonyan:2014ws} and the ResNet~\cite{He:2015tt}. In particular, we use the VGG16 model, which has 16 convolutional layers (configuration C in \cite{Simonyan:2014ws}). For ResNet, we employ the ResNet50 with 50 convolutional layers. To compose our second-order networks, we replace the fully-connected layers and the last average pooling layer with our CDUs. We then attach one fully-connected layer of dimension 23 with softmax activation to obtain the final class probabilities. 
For both VGG and ResNet, to reduce over-fitting, we constrain the weights of the O2T layers to be orthonormal. For the comparison to be fair, and following~\cite{Bell:2015ii}, the weights of the common convolutional layers of both first- and second-order models are initialized with weights pre-trained from ImageNet.  In the following experiments, the CDUs all have 3 O2T layers, with  dimensions set to $D$, $\frac{D}{2}$ and $\frac{D}{4}$, where $D$ is the dimension of the covariance matrix. Note that this does not match the best strategy of Section~\ref{sec:cifar_result}, which consisted of doubling the dimension. However, applying this strategy here would result in a final dimension of $2048$, 

which would significantly increase computational cost. The PV output dimension is the same as that of the last O2T layer.

\prg{Learning Strategy.} 

To train our SO-CNNs (SO-VGG16 and SO-ResNet50), we first freeze the convolutional layers and train the second part of the networks for a few (2-4) epochs, and then fine-tune the whole network in an end-to-end manner. For SO-VGG16, the initial learning rates for second-order training and fine-tuning are set to $10^{-3}$ and $10^{-4}$, respectively, and reduced by a factor 4 when learning plateaued. For SO-ResNet50, the starting rates are set to $10^{-2}$ and $10^{-3}$, respectively.
As mentioned in Section~\ref{sec:transition}, we observed empirically that, during our two-stage learning strategy, we could successfully train the second-order part of the network, but fine-tuning the entire network failed. This, we believe, is due to the fact that, in the first phase, no gradient is backpropagated between the first- and second-order parts of the network. To overcome this, we therefore introduce an additional 1 $\times$ 1 convolutional layer, as described in Section~\ref{sec:transition}. In particular, we set the output dimension of this layer to 512 for SO-VGG16, i.e., the same as the last convolutional layer and to 1024 for SO-ResNet50. 
Note that these models still suffer from the low-rank issue discussed in Section~\ref{sec:robust}. Below, we therefore evaluate on our SO-VGG16 model the effectiveness of our different strategies to addressing this issue, introduced in Section~\ref{sec:robust}. We then compare both SO-VGG16 and SO-ResNet50 to the first-order networks and to the same baselines as in the previous section.

\begin{table}[!t]
\centering
\begin{tabular}{|l|l|l|}
\hline
\textsc{Models}         &  \textsc{Fusion} &  \textsc{Accuracy}                       \\  \hline
$2 \times$ CDU          & V-sum       & $67.86\%$                \\ \hline 
$2 \times$ CDU          & V-avg       & $75.79\%$                \\ \hline 
$2 \times$ CDU          & V-concat    & $75.30\%$               \\ \hline \hline
$4 \times$ CDU          & V-concat    & $74.54\%$  \\ \hline
$8 \times$ CDU          & V-concat    & $76.07\%$  \\ \hline \hline
$2 \times$ CDU          & D-sum       & $75.62\%$                \\ \hline 
$2 \times$ CDU          & D-avg       & $76.42\%$                \\ \hline 
$2 \times$ CDU          & D-concat    & $\boldsymbol{77.88\%}$               \\ \hline \hline
$1 \times$ CDU + Robust & -           & $74.23\%$               \\ \hline 
$2 \times$ CDU + Robust & V-concat    & $76.10\%$               \\ \hline 
$2 \times$ CDU + Robust & D-concat    & $75.17\%$               \\ \hline 
\end{tabular}
\tabcapbefore
\caption{{\bf Comparison of different SO-VGG16 designs.} {\it Robust} indicates the use of a robust covariance estimate, $n \times$ CDU indicates that the convolutional feature maps are split into $n$ groups with one CDU each. V-\texttt{method} indicates that fusion occurs in vector space, while D- stands for descriptor space. 
}
\tabcapafter
\label{tab:minc_cv}
\end{table}

\prg{Robust Estimation \& Fusion of CDUs.}
In Section~\ref{sec:robust}, we introduced two strategies to handle high-dimensional feature maps within our SO-CNNs: Making use of robust covariance estimates and exploiting multiple CDUs. In the latter case, we also proposed several ways to fuse the multiple CDUs into a single representation, consisting of summing, averaging or concatenating the vectors output by the CDUs, or performing similar operations on the final second-order descriptors. We will denote these different fusion strategies by \{V,D\}-sum, \{V,D\}-avg and \{V,D\}-concat, respectively, for the vector (V) or descriptor (D) cases. We report the results of these different strategies in Table~\ref{tab:minc_cv}. These results show that (i) making use of multiple CDUs is typically more effective than relying on a robust covariance estimate; (ii) using more than 2 CDUs has little impact; (iii) fusing at the level of second-order descriptors (D) is more effective than at the level of vectors, particularly via concatenation.

\begin{table}[!t]
\centering
\begin{tabular}{|l|l|l|}
\hline
\multicolumn{3}{|l|}{\textsc{Results on VGG16-based model}} \\
\hline
\textsc{Settings}         &  \textsc{$\#$ Params} &  \textsc{Accuracy}                       \\  \hline
VGG16~\cite{Simonyan:2014ws} & 237M                  & $72.14\%$               \\ \hline 
$1\times 1$ - FCs         & 237.64M               & $70.13\%$               \\ \hline
MatBP~\cite{Ionescu:2015wa}& 20.77M                & $59.06\%$                \\ \hline 
SPD-net~\cite{huang2017}& 0.253M                & $43.90\%$                \\ \hline \hline
Our best                  & 15.21M                & $\boldsymbol{77.88\%}$  \\ \hline
\end{tabular}
\tabcapbefore
\caption{{\bf Baseline comparison on MINC2500 for the VGG16-based models.} We outperform all the baselines significantly, and rely on roughly 90\% fewer parameters than the first-order CNN.}
\tabcapafter
\vspace{-0.1cm}
\label{tab:minc_vgg}
\end{table}

\begin{table}[!t]
\centering
\begin{tabular}{|l|l|l|}
\hline
\multicolumn{3}{|l|}{\textsc{Results on ResNet50-based model}} \\
\hline
\textsc{Settings}         &  \textsc{$\#$ Params} &  \textsc{Accuracy}      \\  \hline
ResNet50~\cite{He:2015tt}       & 23.63M                & $80.10\%$              \\ \hline 
$1\times 1$ - FCs         & 26.17M                & $80.12\%$               \\ \hline
MatBP~\cite{Ionescu:2015wa}& 32.26M                & $55.35\%$                \\ \hline 
SPD-net~\cite{huang2017}& 2.97M                 & $74.33\%$                \\ \hline \hline
Our best                  & 26.00M                & $\boldsymbol{80.45\%}$  \\ \hline
\end{tabular}
\tabcapbefore
\caption{ {\bf Baseline comparison on MINC2500 for the ResNet50-based models.} Note that our SO-ResNet50 again outperforms the second-order-based baselines and the first-order one, although by a smaller margin. We believe that investigating residual second-order strategy could be interesting to further improve our results.
}
\tabcapafter
\vspace{-0.2cm}
\label{tab:minc_resnet}
\end{table}

\prg{Comparison to the Baselines.}
In Tables~\ref{tab:minc_vgg} and~\ref{tab:minc_resnet}, we compare the results of our best SO-VGG16 and the corresponding SO-ResNet50 to the first-order CNNs and to the MatBP~\cite{Ionescu:2015wa} and SPD-net~\cite{huang2017} baselines. Since the SPD-net and MatBP models do not implement any robust covariance  estimation,  we  reduced  the  dimensionality of the feature maps to 512 using  two 1$\times$1 convolutional layers. Without this strategy, the models failed to converge. 
For the comparison with the first-order models to be fair, we also evaluated a version of these models complemented with the same additional 1$\times$ 1 convolutional layer as in our model. The corresponding models are denoted by \emph{ 1 $\times$ 1 - FCs}.
As in the CIFAR-10 case, our end-to-end approach significantly outperforms MatBP and SPD-net, thus showing the benefits of our framework over simpler second-order-based approaches. Our best SO-VGG16 model also outperforms the first-order VGG16 by a significant margin, while relying on much fewer parameters. Note that this is also true for most of the architectures that we have tested in the previous experiment. The fact that the additional 1 $\times$ 1 convolutional layers yields worse accuracy than the original model evidences that the benefit of our method truly comes from the use of second-order information.
For SO-ResNet50, we used the same strategy as for our best SO-VGG16 model. The comparison between our SO-ResNet50 and the first-order ResNet50 model also turns to our advantage. While the margin here is smaller, we believe that many extensions of our model could be studied to make it even more powerful, such as the notion of residual covariances. This, however, will be a topic for future research.

\section{Conclusion}
In this paper, we have introduced an end-to-end learning framework that integrates second-order information for image recognition. To this end, we have developed new layer types and addressed the practical difficulties arising when dealing with covariance matrices. Our experiments have demonstrated that our framework can outperform first-order networks and other second-order-based baselines. In the future, we will explore alternative learning strategies for this type of architectures. We hope that our research will inspire others to investigate architectures that go beyond the standard first-order ones.

\end{document}